\documentclass[10pt,twocolumn,letterpaper]{article}

\usepackage[pagenumbers]{iccv} 
\usepackage{tabularx}
\usepackage{multirow}
\usepackage{graphicx}
\definecolor{iccvblue}{rgb}{0.21,0.49,0.74}
\usepackage[pagebackref,breaklinks,colorlinks,allcolors=iccvblue]{hyperref}

\def\paperID{9642} 
\def\confName{ICCV}
\def\confYear{2025}

\title{DynamicID: Zero-Shot Multi-ID Image Personalization with Flexible Facial Editability}

\author{Xirui Hu$^{1*}$\quad Jiahao Wang$^{1*}$\quad Hao Chen$^2$\quad Weizhan Zhang$^{1\dagger}$\quad Benqi Wang$^2$ \\ Yikun Li$^1$\quad Haishun Nan$^2$ \\
$^1$ Xi'an Jiaotong University  \\
$^2$ Film AI Lab, Western Movie Group \\
}

\begin{document}

\twocolumn[{%
\maketitle
\renewcommand\twocolumn[1][]{#1}%
\begin{center}
    \centering
    \vspace{-22pt}
    \includegraphics[width=\linewidth]{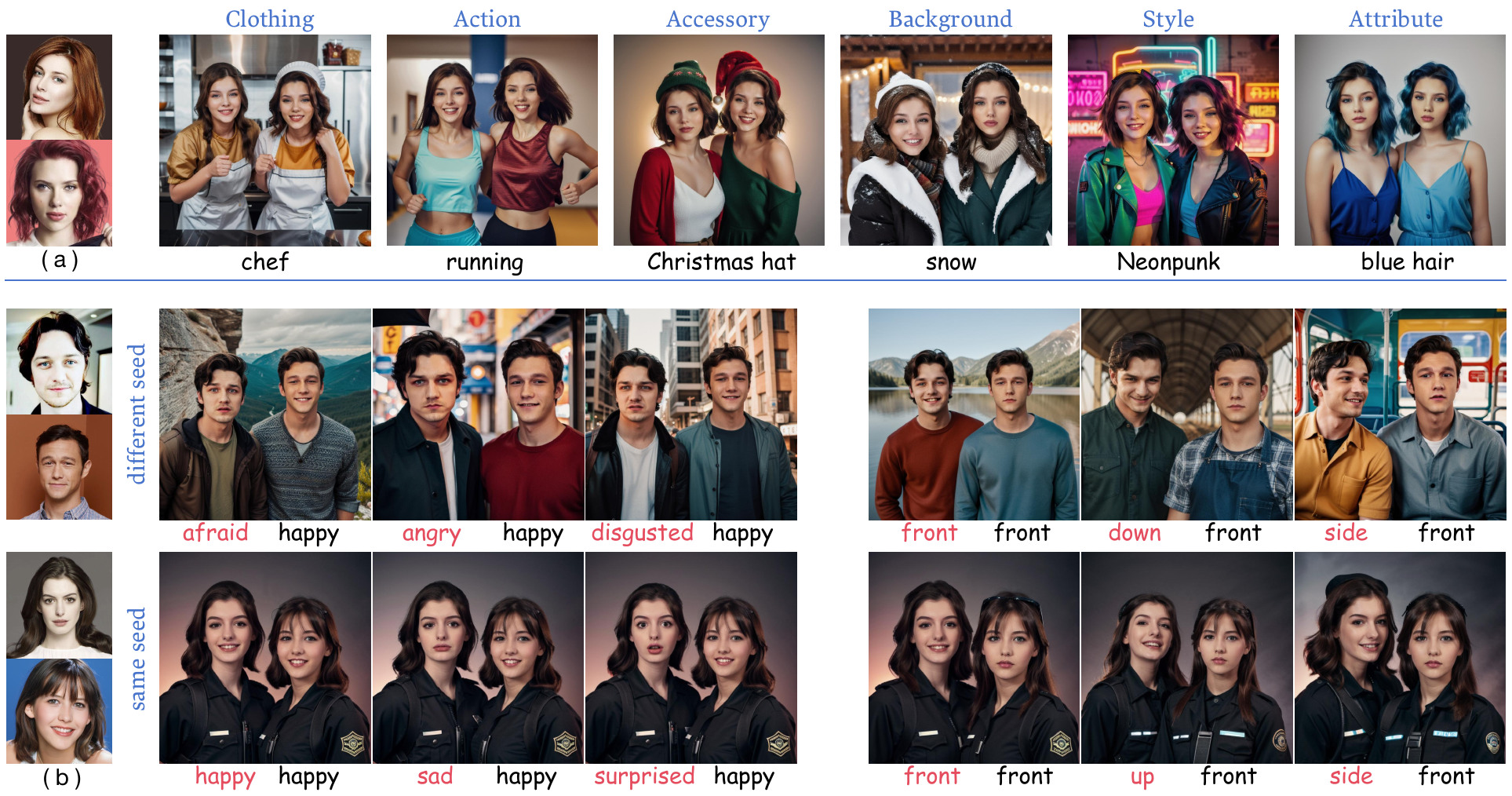}
    \vspace{-20pt}
    \captionof{figure}{Given only a single reference image for each identity, (a) our method demonstrates the capability to generate diverse and personalized multi-ID images based on text prompts. Furthermore, (b) it excels in performing independent and flexible facial editing, especially concerning expressions and orientations, while preserving high fidelity to the original character.}
    \label{fig:teaser}
\end{center}%
}]

\let\thefootnote\relax\footnotetext{$^*$ Equal contributors.\hspace{3pt}$^\dagger$ Corresponding Author}

\begin{abstract}
Recent advances in text-to-image generation have driven interest in generating personalized human images that depict specific identities from reference images. Although existing methods achieve high-fidelity identity preservation, they are generally limited to single-ID scenarios and offer insufficient facial editability. We present DynamicID, a tuning-free framework that inherently facilitates both single-ID and multi-ID personalized generation with high fidelity and flexible facial editability. Our key innovations include: 1) Semantic-Activated Attention (SAA), which employs query-level activation gating to minimize disruption to the base model when injecting ID features and achieve multi-ID personalization without requiring multi-ID samples during training. 2) Identity-Motion Reconfigurator (IMR), which applies feature-space manipulation to effectively disentangle and reconfigure facial motion and identity features, supporting flexible facial editing. 3) a task-decoupled training paradigm that reduces data dependency, together with VariFace-10k, a curated dataset of 10k unique individuals, each represented by 35 distinct facial images. Experimental results demonstrate that DynamicID outperforms state-of-the-art methods in identity fidelity, facial editability, and multi-ID personalization capability.
\end{abstract}    
\section{Introduction}
\label{sec:intro}
In recent years, the field of text-to-image synthesis has seen remarkable progress, largely due to advancements in diffusion-based models \cite{GLIDE, Imagen, StableDiffusion, DALL-E, Flux, SSVQ, Lumina, vividdreamer, infinidreamer}. This progress has spurred interest in personalized human image generation, a domain with extensive applications such as AI-generated portraits, image animation, and e-commerce advertising. The primary goal is to generate images that maintain consistent identities from reference images while seamlessly incorporating additional prompts.

Previous research in personalized human image generation has achieved substantial progress through various approaches \cite{palp, Orthogonal, Apprenticeship, InstructBooth, DisenBooth, DreamArtist, BLIP-Diffusion, ELITE, MagicID2, MindtheGap, HumanDreamer}. Early methodologies \cite{Dreambooth, TextualInversion, HyperDreambooth, CustomDiffusion} demonstrated exceptional high-fidelity outcomes but required subject-specific retraining, imposing significant computational burdens and temporal constraints. To address these limitations, subsequent tuning-free paradigms \cite{CapHuman, InstantID, FlashFace, Face-diffuser, consistentid, FaceStudio} employed pre-trained face encoders coupled with feature injection modules. These frameworks extract facial features through the encoder and subsequently integrate them into the generation process via the injection module. While these parameter-efficient methods eliminate retraining requirements and improve practical deployment capabilities, two critical limitations persist: \textbf{Constraints in Multi-ID Generation}: In scenarios involving multiple identities, numerous approaches struggle with the challenge of identity blending \cite{FastComposer}. While various mitigation strategies have emerged, these methods are built upon frameworks tailored for the single-ID contexts, which inherently compromises model performance. \textbf{Insufficient Facial Attribute Editability}: Current methodologies lack explicit training models to disentangle identity features (\emph{e.g.,} facial structure and skin texture) from motion features (\emph{e.g.,} expression and orientation). This intrinsic entanglement within facial latent representations fundamentally restricts the flexible manipulation of facial attributes while preserving identity fidelity.

To overcome these limitations, we propose \textbf{DynamicID}, a tuning-free method that inherently supports single-ID and multi-ID personalized image generation, ensuring high identity fidelity and flexible facial editability. DynamicID comprises two novel designs: 1) Semantic-Activated Attention (SAA) and 2) Identity-Motion Reconfigurator (IMR). As an innovative feature injection mechanism, the SAA dynamically modulates the activation levels of distinct image latent queries based on their semantic relevance to the facial feature keys. This mechanism ensures that the original model's behavior remains undisturbed during the injection of ID features and allows for free manipulation of activation levels to enable zero-shot layout control and multi-ID personalized image generation. The IMR, as a trainable feature transformer, learns to effectively disentangle and reconfigure identity and motion features of facial latent representations. It serves as an intermediary by capturing a face encoder's output, modifying features, and delivering the modified output to the SAA, thereby enabling high fidelity and flexible facial editability.

In the realm of the training paradigm, end-to-end training serves as a straightforward approach. Yet the demand for large-scale specialized datasets with multi-ID concurrence and free-form facial attributes makes it practically unfeasible. To overcome this challenge, we introduce a task-decoupling training paradigm, characterized by a strategic bifurcation into two distinct stages aimed at separately training the SAA and the IMR. In the initial stage, we concurrently train the SAA and a face encoder, leveraging the innovative mechanism of the SAA, which allows this stage to be conducted using images containing only one individual. Subsequently, our focus shifts to training the IMR which requires only a facial dataset where multiple distinct images represent each individual. To further augment the training of the IMR, we have meticulously curated the VariFace-10k dataset, which encompasses 10k unique individuals, each represented by 35 distinct facial images, thereby providing a flexible and diverse foundation for model enhancement.

The main contributions of this paper are as follows:
\begin{itemize}
    \item We propose DynamicID, a tuning-free framework for personalized image generation that achieves zero-shot adaptation from single-ID to multi-ID contexts while preserving high identity fidelity and flexible facial editability.
    \item We present a Semantic-Activated Attention mechanism, which employs query-level activation gating to refine the feature injection, coupled with an Identity-Motion Reconfigurator module to differentiate facial motion and identity features.
    \item We introduce a task-decoupled training paradigm to streamline model optimization while reducing data dependency, supported by the VariFace-10k dataset comprising 10k identities with 35 systematically varying facial images per identity.
    \item We conduct extensive and comprehensive experiments confirming DynamicID's efficacy and state-of-the-art performance.
\end{itemize}
\section{Related Work}

\subsection{Text-to-Image Models}

 The field of text-to-image generation has experienced significant advancements, driven by innovative techniques such as Generative Adversarial Networks (GANs) \cite{ScalingUpGans, Text-guided-diverse}, auto-regressive models \cite{DALL-E, VisualAutoregressiveModeling, PanoLlama, FutureSightDrive}, and diffusion models \cite{DDPM, StableDiffusion, PRG, Wonderturbo, Muses}. Among these techniques, diffusion models have emerged as the predominant approach, leveraging diffusion processes to iteratively transform noise into coherent images that align with textual descriptions \cite{SDXL, GLIDE, DiT, Kolors, StableDiffusion3, FIX-CLIP}. Our methodology is built upon the stable diffusion model, which is in line with current trends.
 
\subsection{Personalized Human Image Generation}

 Personalized human image generation aims to create customized images that accurately reflect a specific identity based on one or more reference images. Early methods \cite{LoRA, Gradient-free, Vico, Dreamtuner, Cones2} predominantly rely on test-time fine-tuning. For example, DreamBooth \cite{Dreambooth} involved fine-tuning the model parameters on multiple images to learn the target identity. While these methods achieved high fidelity, they demanded substantial computational resources and time. Recent innovative works have introduced novel approaches to address these challenges.

 \textbf{Tuning-free single-ID personalization:} A vast amount of tuning-free studies have concentrated on the single-ID contexts \cite{DemoCaricature, DreamIdentity, ID-Aligner, MagiCapture, Towards-a-Simultaneous, oneactor, DynamicFace}, which typically utilize a face encoder to extract facial features from reference images and an injection module to integrate these features into the generative process. Current approaches can be systematically categorized into three distinct types based on the injection mechanism: text-based methods \cite{face0, Face2Diffusion, PhotoVerse, StableIdentity}, network-based methods \cite{FaceAdapter, FlashFace}, and attention-based methods \cite{Conditioning-diffusion-models, Infinite-ID, LCM-Lookahead, pulid, MasterWeaver}. To illustrate, PhotoMaker \cite{photomaker} extracted facial features and merged them with the corresponding text token within the text embedding space, achieving semantic consistency though compromising identity fidelity. Conversely, InstantID \cite{InstantID} adapts ControlNet \cite{ControlNet} architecture for high-fidelity generation but lacks facial editing granularity. Improving upon this, CapHuman \cite{CapHuman} enhances facial editability via 3D Morphable Face Model \cite{3DMM} integration at the cost of increased computational complexity. As the pioneering attention-based method, IP-Adapter \cite{IP-adapter} introduces a novel decoupled cross-attention mechanism for facial identity integration, achieving superior fidelity while sharing InstantID's limitation in granular facial control. Inspired by this, the $W_+$ adapter \cite{w+adapter} leveraged the StyleGAN \cite{stylegan} encoder to extract features for expression control. However, using this encoder reduced fidelity and limited control over facial orientation. Moreover, these attention-based methods disrupted the model's original behavior during the generative process due to the coarse-grained nature of cross-attention integration \cite{LEGO}.

 \textbf{Tuning-free multi-ID personalization:} Multi-ID contexts remain significantly understudied compared to single-ID contexts, presenting a unique technical challenge: identity blending \cite{FastComposer}, where facial features from different reference subjects intermingle during generation. Recent studies \cite{PotraitBooth, MoA, MagicID, Mix-of-Show, OMG} have proposed several mitigation strategies. Some works build on text-based methods originally tailored for single-ID contexts. For example, FastComposer \cite{FastComposer} addressed identity blending through localized cross-attention maps, while Face-diffuser \cite{Face-diffuser} combined dual specialized diffusion models with Saliency-adaptive Noise Fusion, simultaneously preventing the intermingling of features and enhancing image quality. Other works build on attention-based methods. UniPortrait \cite{Uniportrait} introduced an ID routing module, enabling precise spatial allocation of facial features during synthesis. To prevent feature leakage, InstantFamily \cite{InstantFamily} and MagicID \cite{MagicID} adopted a masked cross-attention mechanism, while StoryMaker \cite{StoryMaker} introduced a cross-attention loss. All three methods incorporated ControlNet to maintain positional correspondence between generated identities and their masked attention maps. Although prior works have addressed identity blending challenges, their reliance on existing frameworks tailored for single-ID context has inadvertently compromised core model functionalities while failing to achieve precise facial editing.

 In this work, we introduce two novel designs: Semantic-Activated Attention (SAA), which could preserve the model's original behavior while inherently supporting both single- and multi-ID contexts, and Identity-Motion Reconfigurator (IMR), which ensures both high fidelity and flexible facial editability.
\section{Methodology}

\begin{figure*}[htbp]
\centering
\includegraphics[width=0.99\textwidth]{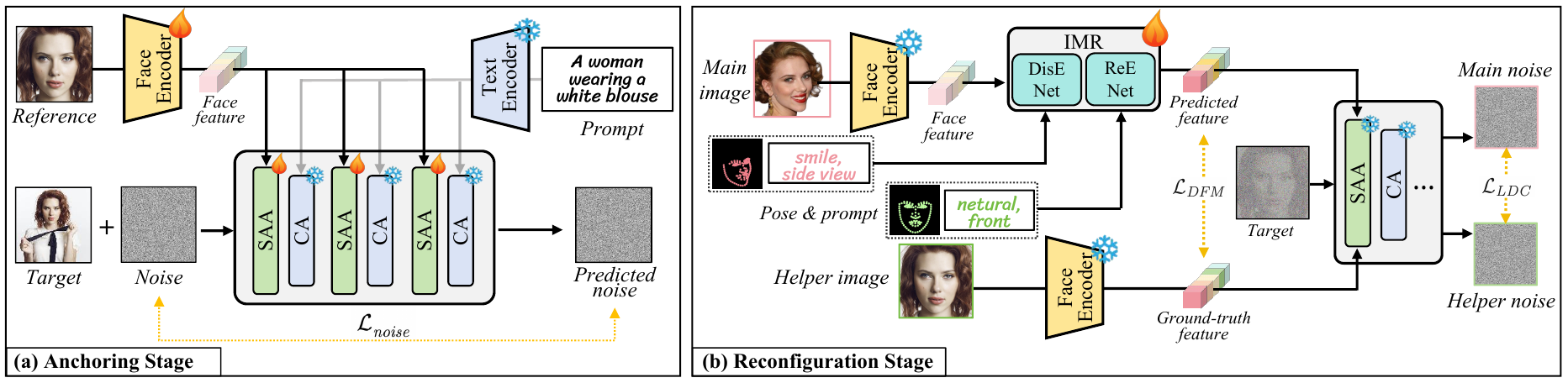}
\caption{\textbf{The training pipeline of the proposed DynamicID.} The proposed framework is architected around two core components: SAA and IMR. (a) In the anchoring stage, we jointly optimize the SAA and a face encoder to establish robust single-ID and multi-ID personalized generation capabilities. (b) Subsequently in the reconfiguration stage, we freeze these optimized components and leverage them to train the IMR for flexible and fine-grained facial editing.}
\label{pipeline}
\end{figure*}

We present a novel unified framework that enables single-ID and multi-ID personalized generation while achieving high identity fidelity and flexible facial editability. Our framework integrates three components: (1) a face encoder extracting facial features from reference images, (2) the Identity-Motion Reconfigurator (IMR) operating in latent space for feature-level editing, and (3) the Semantic-Activated Attention (SAA) effectively injecting processed features into text-to-image models. We introduce the task-decoupled training paradigm structured as anchoring and reconfiguration stages to address the challenge of requiring large-scale specialized datasets. As visualized in Fig.~\ref{pipeline}, the initial anchoring stage establishes fundamental personalized capabilities by simultaneously optimizing the face encoder and the SAA. The subsequent reconfiguration stage specializes in training the IMR for flexible facial editing. Section~\ref{sec:SAA} provides a comprehensive analysis of the SAA alongside implementation details of the anchoring stage. Section~\ref{sec:IMR} systematically elaborates on the IMR architecture and the reconfiguration stage methodology. Furthermore, we contribute the VariFace-10k dataset in Section~\ref{sec:dataset} to support IMR training through diversified facial variations.

\subsection{Semantic-Activated Attention (SAA)}
\label{sec:SAA}
\begin{figure}[h]
\centering
\includegraphics[width=0.47\textwidth]{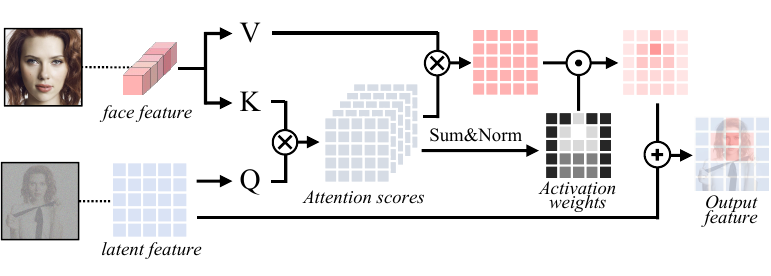}
\caption{\textbf{The mechanism of the proposed Semantic-Activated Attention.} The SAA mitigates the fixed attention mass constraint of standard cross-attention by adaptively modulating the activation weights of the queries.}
\label{SAA_pipeline}
\end{figure}

The predominant technique for feature injection in personalized generation is the cross-attention mechanism. However, the inherent softmax function introduces an implicit inductive bias that forces each query to distribute a fixed total attention mass across keys, even in the absence of semantic relevance between the query and any of the keys. This behavior can lead to two significant issues: 1) perturbations to the model's original behavior, and 2) identity blending \cite{FastComposer} in multi-ID personalization. To overcome these limitations, we propose Semantic-Activated Attention (SAA). As shown in Fig.~\ref{SAA_pipeline}, given the latent image representation $z$ and facial features $c_f$ extracted from the reference image, SAA is formally defined through the following operations:
\begin{align}
    z_{\textit{new}} = z + \mathrm{Expand}&(w) \odot \mathrm{softmax}\left(\frac{QK^\top}{\sqrt{d}}\right) V, \\
    w = &\mathrm{Norm}(QK^\top J),
\end{align}


where $Q = zW_q$, $K = c_fW_k$, and $V = c_fW_v$ represent the attention mechanism's query, key, and value matrices, respectively. The matrix $J \in \mathbb{R}^{k \times 1}$ denotes an all-ones column vector $\begin{bmatrix} 1 & 1 & \cdots & 1 \end{bmatrix}^\top$ with length matching the number of keys, $k$. The $\mathrm{Expand}(\cdot)$ operator broadcasts the weight vector $w$ along the feature dimension to match the dimension of the attention output matrix. The $\mathrm{Norm}(\cdot)$ operator applies min-max normalization to scale its input values to the [0, 1] range. The activation weights, derived from the interactions between queries and keys, effectively indicate the extent to which each query should be associated with the reference facial information, as verified in Fig.~\ref{activate_weight}. Intuitively, SAA enhances ID features integration by ensuring that queries corresponding to the facial region receive strong activation (with their corresponding activation weight approaching 1) to incorporate reference facial information, while queries corresponding to the background are suppressed (with their corresponding activation weight approaching 0) according to their inherent irrelevance to the reference face. Additionally, queries corresponding to the body are moderately activated to maintain the internal coherence of the image.

The incorporation of a query-level activation gating mechanism confers inherent zero-shot generalization properties upon our pipeline, manifesting through three principal dimensions:


\textbf{Context Decoupling:} The novel SAA suppresses queries originating from non-facial regions during inference, ensuring that the generation process in these areas remains unaffected by extraneous facial information. As a result, the spatial content and motion dynamics of non-facial regions remain consistent when the reference identity changes, as shown in Fig.~\ref{SAA_ablation}(a).

\textbf{Layout Control:} The novel SAA enables parametric modulation of activation weights during early denoising steps to achieve layout control as shown in Fig.~\ref{SAA_ablation}(b). This controlled intervention, scaled by parameter $\beta$, induces the semantic transition of targeted regions toward facial features. Global coherence is maintained through multi-level query activation in subsequent denoising stages, demonstrating superiority over binary activation approaches. The intervention is formally expressed as:
\begin{align}
\hat w = M \odot(w+\beta)+(1-M)\odot(\frac{w}{\beta+1}),
\end{align}
where $M$ represents the user-specified mask region, with the hyperparameters $\alpha$ and $\beta>1$ governing the temporal scope and intervention intensity, respectively.

\textbf{Multi-ID Personalization:} The layout control capability of SAA is instrumental in multi-ID contexts, enabling the spatial arrangement of each character's face to be predetermined before inference. Specifically, during the incorporation of facial information for a particular character, queries associated with facial generation regions of other characters can be selectively suppressed to prevent identity blending. This approach facilitates multi-ID personalization generation without necessitating additional modules or training, as shown in Fig.~\ref{SAA_ablation}(b).

Notably, besides its versatility, the SAA framework maintains remarkable training efficiency, requiring only the single-ID dataset which is readily accessible. For the face encoder, we employ two pre-trained feature extractors to independently capture global and local facial features, followed by a projection module synthesizing these features into a unified facial latent representation. In the anchoring stage, we freeze the parameters of the pre-trained extractors, training only the projection module and SAA, using the same training objective as the original diffusion process:
\begin{align}
\mathcal{L}_\textit{noise} = E_{z,t,\xi,\tau,\epsilon}||\epsilon - \epsilon_{\theta}(z_t,t,\xi,\tau)||_2^2,
\end{align}
 where $\xi$	and $\tau$ represent the reference facial features extracted by the face encoder and the textual features extracted by the text encoder, respectively, while $\epsilon_{\theta}$ represents the T2I model incorporating the SAA.

\subsection{Identity-Motion Reconfigurator (IMR)}
\label{sec:IMR}
To achieve personalized generation with flexible and fine-grained facial editability, we introduce the Identity-Motion Reconfigurator (IMR) within the interaction space of the face encoder and the SAA. The IMR comprises two components: DisentangleNet $\boldsymbol{\phi_1}$ and EntangleNet $\boldsymbol{\phi_2}$, which strategically utilize face prompts to extract facial motion characteristics and incorporate subtle perturbed facial landmarks to improve the accuracy of the estimation of the orientation angle. Face prompts are encoded into $f_p \in \mathbb{R}^{1 \times d_p}$ through a text encoder, while landmarks are transformed into $f_l \in \mathbb{R}^{1 \times d_l}$ via a keypoint encoder. These representations are subsequently projected into $c$ tokens using a Multi-layer Perceptron (MLP), expressed as:
\begin{align}
\psi = \textit{MLP}(\mathrm{Concat}(f_l,f_p) ) \in \mathbb{R}^{c \times d}.
\end{align}

DisentangleNet, using the facial features \(\xi_{\textit{src}}\) extracted by the face encoder trained during the anchoring stage, together with the corresponding motion features \(\psi_{\textit{src}}\), predicts the posterior distribution of the identity features \(p(\zeta | \xi_{\textit{src}}, \psi_{\textit{src}})\). Subsequently, EntangleNet forecasts the distribution of the target facial features \(p(\xi_{\textit{pred}})\) by integrating the identity features \(\zeta\) with the target motion features \(\psi_{\textit{tgt}}\).
\begin{align}
\xi_\textit{pred} = \boldsymbol{\phi_2}(\boldsymbol{\phi_1}(\xi_{\textit{src}},\psi_{\textit{src}}),\psi_{\textit{tgt}}).
\end{align}

Consequently, by disentangling and reconfiguring facial features, the IMR captures the feature output of the face encoder, effectively modifies its motion features towards the desired direction, and then provides these modified features to the SAA. Thereby, DynamicID facilitates fine-grained and flexible control of facial motion while preserving identity fidelity.

During inference, the IMR effortlessly handles precise and detailed face prompts and remains effective even when prompts are reduced to two fundamental attributes: expression and orientation. These attributes can be reliably decomposed from input faces using lightweight pre-trained models. Furthermore, incorporating subtly perturbed facial landmarks during training eliminates the need for exact source and target landmarks.

Thanks to the innovative design of our IMR, which functions within latent representations, its training protocol only requires a dataset with various facial images of each individual. During the training phase, we commence by selecting a collection of $m$ facial images of a specific individual, denoted as $\{x^i\}_{i=1}^{m}$. Subsequently, we extract the corresponding motion features of face prompts and landmarks, $\{\psi^i\}_{i=1}^{m}$. Leveraging the face encoder, we project the original images into the feature space, yielding $\{\xi^i\}_{i=1}^{m}$. In each training iteration, a single feature is randomly designated as the source feature, while the remaining features function as the target features. The IMR is then trained to employ motion cues to morph the source feature into the target feature, guided by our dual-objective loss function:
\begin{align}
\mathcal{L}_{\textit{edit}} = \underbrace{||\xi_{\textit{pred}} - \xi_{\textit{tgt}}||^2_2}_{\textit{Direct Feature Matching}} + \lambda \underbrace{||\epsilon_{\theta}^{\prime}(\xi_{\textit{pred}}) - \epsilon_{\theta}^{\prime}(\xi_{\textit{tgt}})||^2_2}_{\textit{Latent Diffusion Consistency}},
\end{align}
where $\epsilon_{\theta}^{\prime}$ denotes the optimized version of the original model $\epsilon_{\theta}$ after the anchoring stage. The Direct Feature Matching term enforces that the predicted representation aligns with the target representation in the explicit feature space, preserving critical attributes through numerical proximity. Concurrently, the Latent Diffusion Consistency term guarantees semantic equivalence in the T2I model’s implicit space, ensuring predicted features induce similar generative behaviors as target features.

\subsection{VariFace-10k Dataset}
\label{sec:dataset}

Our IMR requires a comprehensive personalized dataset consisting of various facial images for each individual, showcasing different expressions, orientations, and other attributes, paired with corresponding prompts. However, existing high-quality facial datasets, such as FFHQ \cite{FFHQ}, SFHQ \cite{SFHQ}, and CelebA \cite{CelebA}, lack sufficient diversity in images of the same individual. To address this issue, we constructed the VariFace-10k dataset, where each individual is represented by 35 distinct facial images exhibiting significant variations, essential for mitigating the current scarcity of personalized datasets. More details about the VariFace-10k dataset are available in the appendix.

\begin{figure}[hbtp]
\centering
\includegraphics[width=0.46\textwidth]{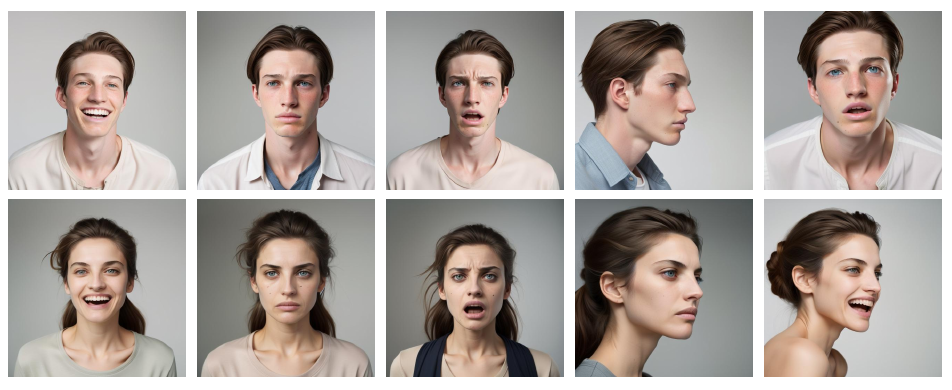}
\caption{\textbf{An illustration of selected samples from our constructed VarFace-10k dataset.} The VarFace-10k dataset contains multiple facial images per individual, exhibiting significant variations in facial expressions, head poses, lighting conditions, and other attributes.}
\label{dataset}
\end{figure}
\section{Experiment}
\subsection{Experimental Setup}
\textbf{Implementation details.} We employ the SD1.5 model \cite{StableDiffusion} as the foundational text-to-image model. For the face encoder, we utilize buffalo\_l \cite{ArcFace} as the global feature extractor and CLIP-ViT-H \cite{laion-5b} as the local feature extractor. Within the IMR architecture, the DisentangleNet and EntangleNet modules are identically structured, each comprising a single cross-attention layer followed by a self-attention layer and an MLP layer. We utilize clip-vit-large-patch14 \cite{CLIP} as the text encoder and a simple CNN \cite{CNN} as the keypoint encoder. For the anchoring training stage, we filter 10k high-quality human images from the Laion-Face \cite{laion-face} dataset. For the reconfiguration training stage, we utilize the VariFace-10k dataset (with a 95\% probability) and the KDEF \cite{KDEF} dataset (with a 5\% probability). Additionally, we configured the hyperparameter $\lambda$ to 1. All training is conducted on 8 NVIDIA A100 GPUs, utilizing the AdamW \cite{AdamW} optimizer with a batch size of 16 and a constant learning rate of 1e-5. During inference, we employ 50-step DDIM \cite{DDIM} sampling with a classifier-free guidance scale of 5. Detailed implementation of the Latent Diffusion Consistency term is provided in the appendix.

\begin{figure*}[htbp]
\centering
\includegraphics[width=0.9\textwidth]{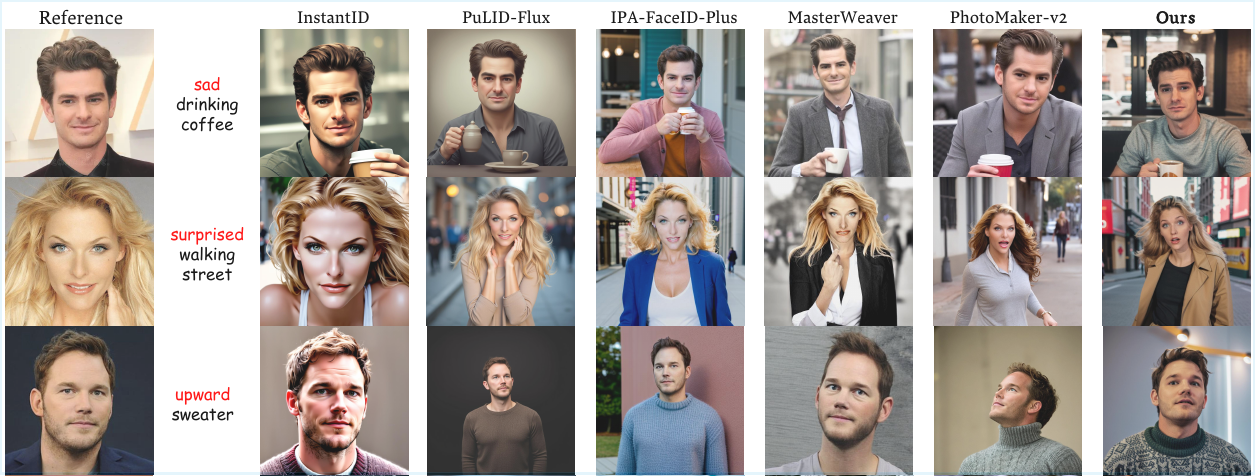}
\caption{\textbf{Visual Comparison on Single-ID Personalization.} All images are generated using the single reference image shown on the left. Our method uniquely achieves flexible and fine-grained editability of facial features, producing high-quality images with faithful identity preservation. Zoom in for a better view.}
\label{qualitative single-id}
\end{figure*}

\textbf{Evaluation details.} Following established evaluation protocols, we adopt the CLIP-T \cite{CLIP} metric for text-image alignment assessment and the FaceSim \cite{FaceNet} metric for facial identity preservation. To facilitate the assessment of facial editability, we introduce two novel metrics: 1) \textit{Expr}: This metric quantifies the effectiveness of expression editing by inputting the generated image into an expression classification model \cite{ResEmoteNet} and extracting the probability corresponding to the specified expression. 2) \textit{Pose}: This metric evaluates viewpoint control accuracy using a head orientation estimation model \cite{ArcFace} to categorize synthesized faces into four distinct viewpoint categories (front, side, up, and down) based on yaw and pitch angle thresholds. We employ an automated face detection pipeline with per-subject metric aggregation in the multi-ID contexts. For quantitative analysis, we randomly select identities from the CelebA \cite{CelebA} dataset to form our test set, following \cite{FastComposer}. We augment prompts with explicit descriptors for expression and orientation to assess the model's facial editability. Refer to the appendix for further details.

\begin{figure*}[htbp]
\centering
\includegraphics[width=0.90\textwidth]{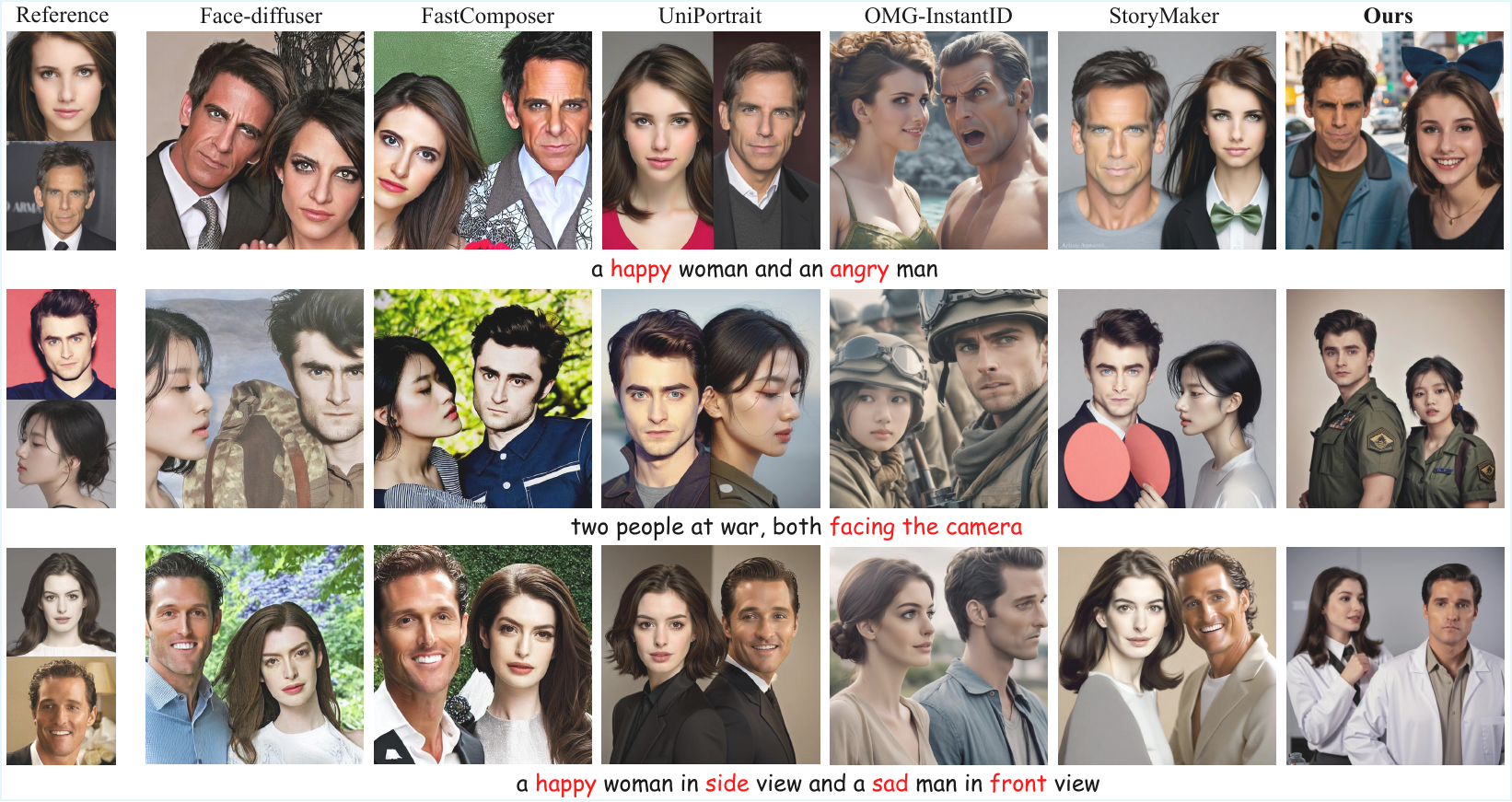}
\caption{\textbf{Visual comparison on Multi-ID Personalization.} Current methods demonstrate constraints in multi-ID generation and unsatisfying editability. In contrast, our method enables flexible facial editability and high image quality in such scenarios.}
\label{qualitative multi-id}
\end{figure*}

\subsection{Comparation Results}
\textbf{Single-ID personalization.} To demonstrate the effectiveness of our DynamicID for single-ID personalization, we conducted extensive comparative experiments against state-of-the-art baselines, including \cite{InstantID,pulid, IP-adapter, MasterWeaver,photomaker}. The quantitative results are presented in Table \ref{quantitative single-id}, while the qualitative outcomes are illustrated in Fig.\ref{qualitative single-id}. Our analysis reveals that while IPA-FaceID-Plus and InstantID demonstrate high fidelity, they exhibit significant limitations in facial editability. The high FaceSim scores for these methods are primarily due to their tendency to direct facial replication, as evidenced by their low Expr and Pose scores. PhotoMaker-v2 offers facial editing capabilities but compromises on fidelity. MasterWeaver attempts to balance expression editing and facial fidelity, yet fails to achieve satisfactory results. Despite being built upon the highly parameter-intensive FLUX model \cite{Flux}, PuLID fails to deliver facial editing capabilities while significantly increasing resource consumption. In contrast, our proposed method strictly adheres to the prompts, generating images with flexible facial editability while maintaining strong identity fidelity. The visual comparison also intuitively highlights our method's ability to generate high-quality images across diverse domains, including clothing, backgrounds, actions, and styles. Additional qualitative results, including complex expression editing, layout control, context decoupling, and ID mixing, are provided in the appendix.

\begin{table}[ht]
  \centering
  \resizebox{0.98\linewidth}{!}{
  \begin{tabular}{lccccc}
    \toprule[1.5pt]
    Method  & Arch. & CLIP-T $\uparrow$ & FaceSim $\uparrow$ & Expr $\uparrow$ & Pose $\uparrow$ \\
    \midrule
    PuLID-FLUX & FLUX & 0.237 & 0.667 & 0.181 & 0.273 \\
    PhotoMaker-v2 & SDXL & 0.238 & 0.592 & 0.243 & 0.869 \\
    InstantID  & SDXL & 0.233 & \textbf{0.723} & 0.151 & 0.264 \\
    IPA-FaceID-Plus & SD1.5 & 0.236 & 0.712 & 0.156 & 0.266 \\
    MasterWeaver & SD1.5 & 0.237 & 0.651 & 0.189 & 0.278 \\
    Ours & SD1.5 & \textbf{0.239} & 0.671 & \textbf{0.456} & \textbf{0.878} \\
    \bottomrule[1pt]
  \end{tabular}}
  \caption{\textbf{Quantitative comparison between our DynamicID and state-of-the-art methods on single-ID Personalization.}}
  \label{quantitative single-id}
\end{table}

\begin{figure}[htbp]
\centering
\includegraphics[width=0.47\textwidth]{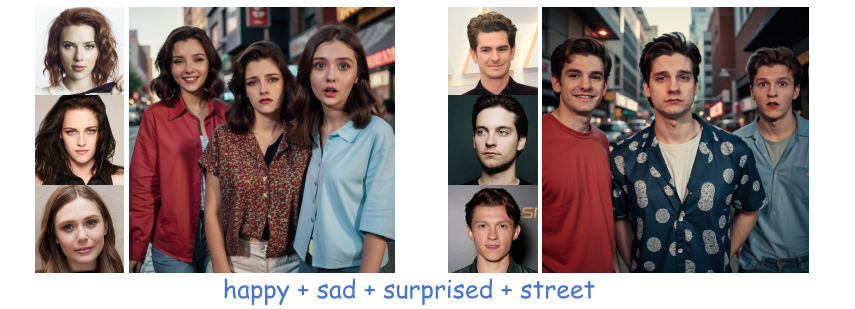}
\caption{\textbf{Illustration of three IDs personalization.} Our DynamicID framework is capable of effectively handling scenarios involving more than two individuals, while simultaneously enabling independent facial editing for each person.}
\label{moreid}
\end{figure}

\textbf{Multi-ID personalization.} To validate the efficacy of our DynamicID for multi-ID personalization, we conducted comprehensive comparisons with existing state-of-the-art methods, including those proposed in \cite{FastComposer, Face-diffuser, Uniportrait, OMG, StoryMaker}. Figure \ref{qualitative multi-id} showcases a visual comparison of qualitative results, while Table \ref{quantitative multi-id} provides the quantitative analysis. These evaluations underscore that our method uniquely enables independent, fine-grained editing of facial features. In contrast, FastComposer, Face-diffuser, Uniportrait, OMG-InstantID and StoryMaker face challenges with direct facial replication. Furthermore, while maintaining comparable prompt consistency with baseline approaches, our method significantly enhances the visual quality of generated images. The higher score of Uniportrait and OMG-InstantID in the FaceSim metric compared to our method stems from the same reason as discussed in the context of single-ID personalization. More visual examples for three-ID images are shown in Fig.~\ref{moreid}.

\begin{table}[ht]
  \centering
  \resizebox{\linewidth}{!}{
  \begin{tabular}{lccccc}
    \toprule[1.5pt]
    Method & Arch. & CLIP-T $\uparrow$ & FaceSim $\uparrow$ & Expr $\uparrow$ & Pose $\uparrow$ \\
    \midrule
    FastComposer & SD1.5 & 0.233 & 0.594 & 0.144 & 0.256 \\
    
    Face-Diffuser & SD1.5 & 0.234 & 0.612 & 0.149 & 0.264 \\

    UniPortrait & SD1.5 & 0.235 & \textbf{0.718} & 0.149 & 0.268 \\

    OMG-InstantID  & SDXL & 0.231 & 0.657 & 0.343 & 0.742 \\
    
    StoryMaker  & SDXL & 0.219 & 0.678 & 0.147 & 0.296 \\

    Ours & SD1.5 & \textbf{0.237} & 0.664 & \textbf{0.431} & \textbf{0.867} \\
    \bottomrule[1pt]
  \end{tabular}}
  \caption{\textbf{Quantitative comparison between our DynamicID and state-of-the-art methods on multi-ID Personalization.}}
  \label{quantitative multi-id}
\end{table}

\subsection{Ablation study}
\begin{figure}[htbp]
\centering
\includegraphics[width=0.47\textwidth]{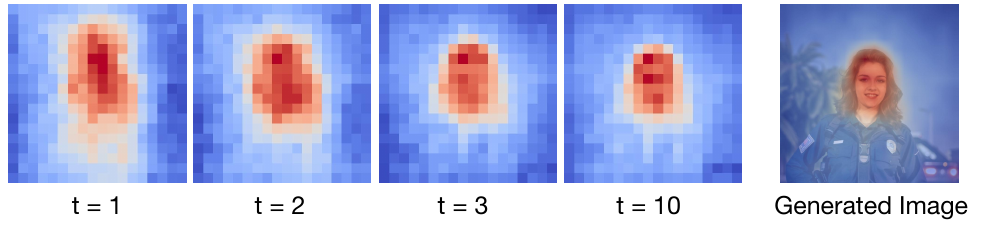}
\caption{\textbf{Query activation levels across varying timesteps.}}
\label{activate_weight}
\end{figure}

\textbf{Motivation Verification:} Fig.~\ref{activate_weight} shows that during the entire denoising process, only the queries corresponding to facial regions exhibit strong activation. In contrast, queries corresponding to background regions remain non-activated, thereby preserving the behavior of the T2I model in these regions without interference.

\begin{figure}[htbp]
\centering
\includegraphics[width=0.47\textwidth]{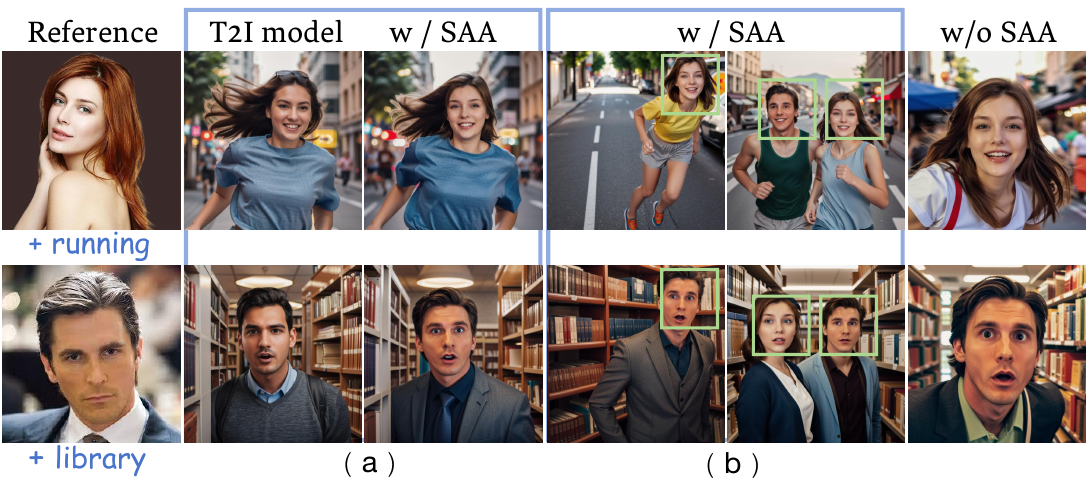}
\caption{\textbf{Ablation study on SAA.} (a) The SAA preserves the original model's behavior when queries are adaptively activated. (b) Moreover, it allows for the manual enhancement of query activation levels within user-defined spatial regions via bounding box annotations, thereby enabling zero-shot layout control and multi-ID personalized generation.}
\label{SAA_ablation}
\end{figure}

\textbf{Impact of SAA:} Fig.~\ref{SAA_ablation} illustrates the effectiveness of the SAA. With the integration of SAA, the model's behavior aligns more closely with that of the base T2I model, thereby preserving the robust text-editing capabilities inherent to the base T2I model. Moreover, SAA allows for manual adjustment of the query activation levels, which enables zero-shot layout control and multi-ID personalized generation. As presented in Table \ref{ablation study}, quantitative experiments show a reduction in the CLIP-T score without SAA, indicating a decrease in the model's text-editing proficiency. An extended analysis of the parametric sensitivity of \(\alpha\) and \(\beta\) on multi-ID personalization, is presented in the appendix.

\begin{figure}[htbp]
\centering
\includegraphics[width=0.47\textwidth]{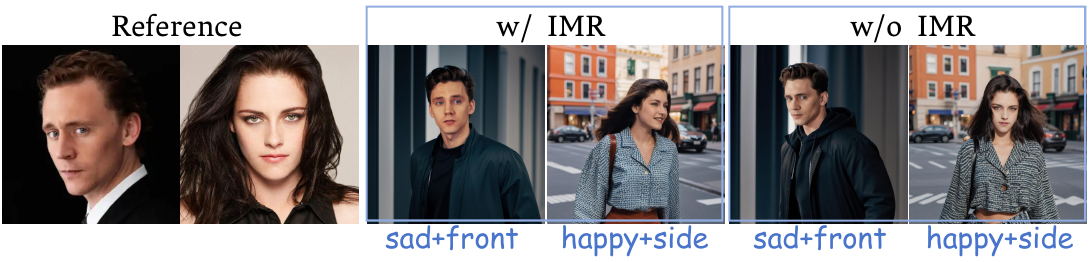}
\caption{\textbf{Ablation study on IMR.} The IMR plays a significant role in enabling flexible and fine-grained facial editing.}
\label{IMR_ablation}
\end{figure}

\begin{table}[ht]
  \centering
  \resizebox{\linewidth}{!}{
  \begin{tabular}{lcccc}
    \toprule[1.5pt]
    Method & CLIP-T $\uparrow$ & FaceSim $\uparrow$ & Expr $\uparrow$ & Pose $\uparrow$ \\
    \midrule
    Ours w/o SAA & 0.224 & 0.682 & 0.422 & 0.862 \\

    Ours w/o IMR & 0.228 & \textbf{0.712} & 0.161 & 0.253 \\

    Ours & \textbf{0.239} & 0.671 & \textbf{0.456} & \textbf{0.878} \\
    \bottomrule[1pt]
  \end{tabular}}
  \caption{\textbf{Ablation study on SAA and IMR.}}
  \label{ablation study}
\end{table}

\textbf{Impact of IMR:} Table \ref{ablation study} presents ablation study on our proposed IMR. Without IMR, the model exhibits significant limitations in flexible facial editing, as reflected in the degraded Expr and Pose metrics. The increased FaceSim score stems from the model replicating the reference face rather than generating meaningful variations. These findings are visually supported by Fig.~\ref{IMR_ablation}. Additional ablation experiments on the Direct Feature Matching and Latent Diffusion Consistency terms are available in the appendix.

\section{Conclusion}
We present DynamicID, a novel framework that enables high identity fidelity and flexible facial editing in personalized image generation across single- and multi-ID scenarios. Our approach introduces two novel components: 
SAA and IMR
and employs task-decoupled training to eliminate dependency on specialized datasets, complemented by our newly developed VariFace-10k dataset containing diverse facial variations across 10k identities. Extensive experiments demonstrate state-of-the-art performance, validating the method's effectiveness and broad application potential in personalized generation.

\section*{Acknowledgement}
This work was supported in part by the National Natural Science Foundation of China under Grant 62192781, 62172326, 62137002, and in part by the Project of China Knowledge Centre for Engineering Science and Technology.

\bibliographystyle{ieeenat_fullname}
{
    \small
    \bibliography{main}
}
\clearpage
\section{Appendix}
\label{sec:apdix}

\subsection{Limitation and Social Impact}
Although our plug-and-play framework fundamentally supports all text-to-image models, this work primarily uses Stable Diffusion v1.5 \cite{StableDiffusion}, which limits performance when generating faces near image edges \cite{spotactor} or handling complex prompts \cite{Agentthink, Priormotion}. Future work will explore scaling via larger datasets \cite{DMM, DIM,SMVTEP, PrismLayers, Bizgen}, broader model adaptation, and version releases. Our method's ability to generate realistic images of individuals raises ethical concerns, particularly for deepfakes \cite{mllmattack}. We will explore safeguards like digital watermarks to mitigate this risk.

\subsection{Detail process of SAA}
\begin{figure*}[h]
  \centering
  \includegraphics[width=1\linewidth]{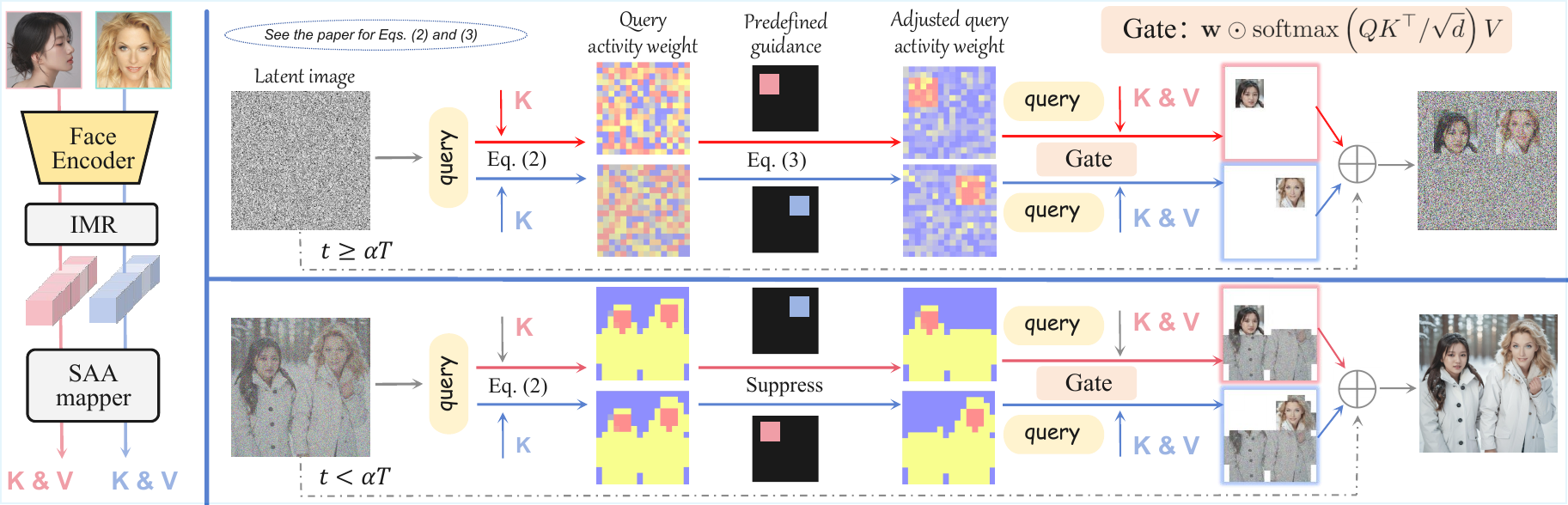}
   \caption{
   \small
   Detail process of SAA in multi-ID personalization.}
   \label{SAAinMulti}
\end{figure*}

A diagram of the process of SAA in multi-ID personalization is provided in Fig.~\ref{SAAinMulti}.

\subsection{VariFace-10k dataset details}
The training of our IMR necessitates a comprehensive personalized dataset, where each identity is characterized by a diverse collection of facial images exhibiting a wide spectrum of expressions, orientations, and other attributes, complemented by corresponding textual prompts. This comprehensive dataset is crucial for advancing the model's understanding of the disentanglement and entanglement between identity and motion features within the feature space. However, currently, available high-quality facial datasets, including FFHQ \cite{FFHQ}, SFHQ \cite{SFHQ}, and CelebA \cite{CelebA}, demonstrate significant limitations in terms of intra-individual image diversity, typically constrained to a narrow range of expressions (predominantly neutral and happy) and even representing individuals with only a single image. To overcome these limitations, we have developed the VariFace-10k dataset, which contains 35 distinct facial images per individual, each exhibiting substantial variations across multiple dimensions. This dataset serves as a fundamental resource for training our IMR and addresses the existing gap in personalized dataset availability. Our dataset construction process involved initially curating high-quality facial images from the FFHQ dataset, subsequently augmenting this collection with additional high-quality images sourced from the internet and further expanding the dataset through GAN-based generation of supplementary high-quality facial images. All images were standardized through uniform cropping to 512x512 resolution, resulting in a foundational set of 10k distinct facial images. Building upon this foundation, we employed the Face-Adapter \cite{FaceAdaptor} to perform face reenactment using images from the KDEF dataset as driving images, ultimately generating an extensive collection of 350k facial images comprising 10k unique identities, each represented by 35 distinct facial attributes. Recognizing the potential for facial distortion in generating profile views from frontal images, we processed each set of 35 images per individual through the IP-Adapter-FaceID-Portrait model \cite{IP-adapter} to regenerate 35 refined images. Finally, we implemented \cite{ArcFace} for landmark generation and utilized \cite{InternVL} to provide detailed, fine-grained textual prompts for each facial image in our dataset.

\subsection{More evaluation details}
For quantitative analysis, we randomly selected 500 identities from the CelebA dataset to construct our test set, adhering to the methodology outlined in \cite{FastComposer}. We employed 20 prompts encompassing various accessories, clothing, backgrounds, actions, and styles. Table \ref{tab:prompta} provides the complete list of prompts. Each base prompt was systematically augmented through the injection of supplementary facial attributes, including seven distinct facial expressions (neutral, happy, angry, disgusted, surprised, sad, afraid) and four orientation descriptors (front view, side view, facing up, facing down). Single-ID prompts are structured as: a person with a \textit{happy} expression \textit{in a side view}, \textit{wearing headphones}. Multi-ID prompts are: The person on the left has a neutral expression in a side view, and the person on the right has a \textit{sad} expression \textit{in a front view}, both \textit{wearing headphones}. For the \textit{Pose} metric, if the source prompt includes "facing up" or "facing down," we utilize the pitch angle with a threshold of 10 degrees to categorize the images into 'up,' 'down,' or 'front' classes. Conversely, if the source prompt contains "in front view" or "in side view," we employ the yaw angle with a threshold of 10 degrees to classify the images into 'side' or 'front' categories.

\subsection{Detailed implementation of LDC term}
The \textbf{L}atent \textbf{D}iffusion \textbf{C}onsistency term can be more precisely expressed as:
\begin{align}
||\epsilon_{\theta}(z_t, t, \xi_{pred}, \tau) - \epsilon_{\theta}(z_t, t, \xi_{tgt}, \tau)||_2^2
\end{align}
where \(z_t\) is derived from the target facial image, and \(\tau\) is the text embedding corresponding to the facial prompt associated with the target image. Within this framework, the Latent Diffusion Consistency term ensures semantic equivalence in the T2I model's latent space, ensuring that the predicted features induce generative behaviors similar to those of the target features.

\subsection{More ablation}
\begin{figure}[ht]
\centering
\includegraphics[width=0.46\textwidth]{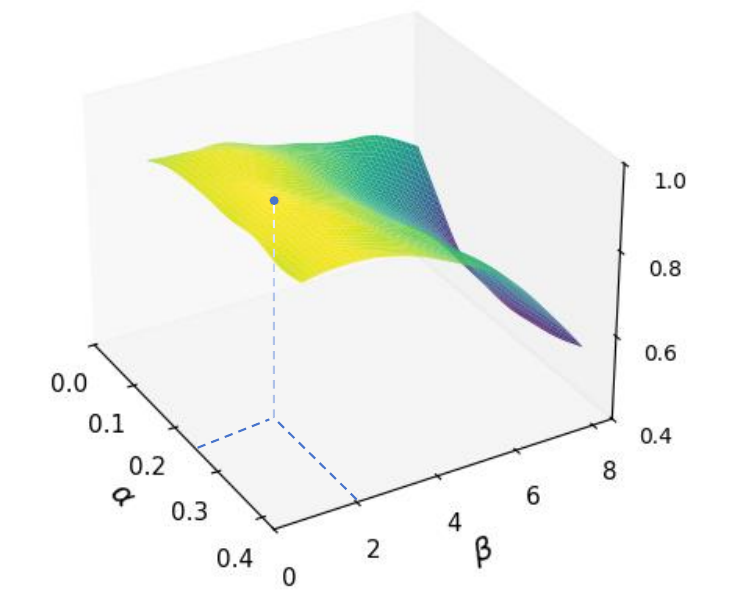}
\caption{Investigating the impact of varying hyperparameter values of \(\alpha\) and \(\beta\) for multi-ID personalized generation, our experimental results led us to select \(\alpha = 0.24\) and \(\beta = 2\).}
\label{3d}
\end{figure}

\textbf{Effect of parameters $\alpha$ and $\beta$ on multi-ID personalized generation:} We conducted an in-depth investigation into the effects of parameters $\alpha$ and $\beta$ on multi-ID personalized generation, using the probability of detecting valid faces across all target regions as the evaluation metric. As illustrated in Fig.~\ref{3d}, the experimental results led us to select $\alpha=0.24$ and $\beta=2$ as the optimal values.

\textbf{Effect of the two Terms in the IMR training stage:} As evidenced in Table \ref{ablation term study}, both the Direct Feature Matching term and the Latent Diffusion Consistency term play pivotal roles in attaining flexible facial editability while maintaining high identity preservation. Our ablation study demonstrates that the elimination of the Latent Diffusion Consistency term substantially impairs facial editing capability, whereas the removal of the Direct Feature Matching term significantly compromises identity fidelity. These empirical findings underscore the complementary nature and synergistic interplay of these two terms in achieving optimal performance in the personalized generation.

\begin{table}[h]
  \centering
  \begin{tabular}{lcccc}
    \toprule
    \textbf{Method} & \textbf{CLIP-T}  & \textbf{FaceSim} & \textbf{Expr} & \textbf{Pose}  \\
    \midrule
    Ours w/o DFM & 0.237 & 0.663 & 0.433 & 0.851 \\

    Ours w/o LDC & 0.238 & 0.667 & 0.234 & 0.644 \\

    Ours & \textbf{0.239} & \textbf{0.671} & \textbf{0.456} & \textbf{0.878} \\
    \bottomrule
  \end{tabular}
  \caption{The proposed \textbf{D}irect \textbf{F}eature \textbf{M}atching term and \textbf{L}atent \textbf{D}iffusion \textbf{C}onsistency term significantly enhance flexible facial editability and maintain identity fidelity.}
  \label{ablation term study}
\end{table}

\subsection{More Applications}
\begin{figure*}[b]
\centering
\includegraphics[width=0.98\textwidth]{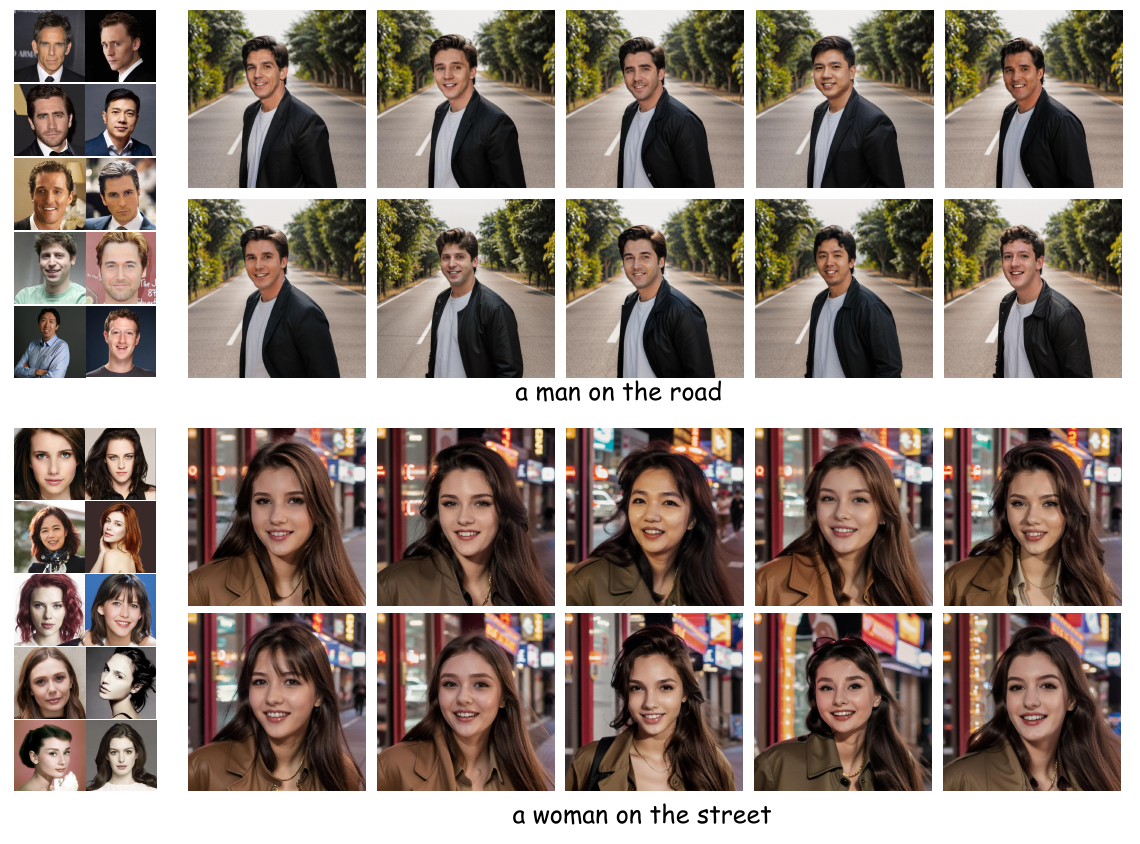}
\caption{The application of context decoupling.}
\label{context}
\end{figure*}

\begin{figure*}[h]
\centering
\includegraphics[width=0.98\textwidth]{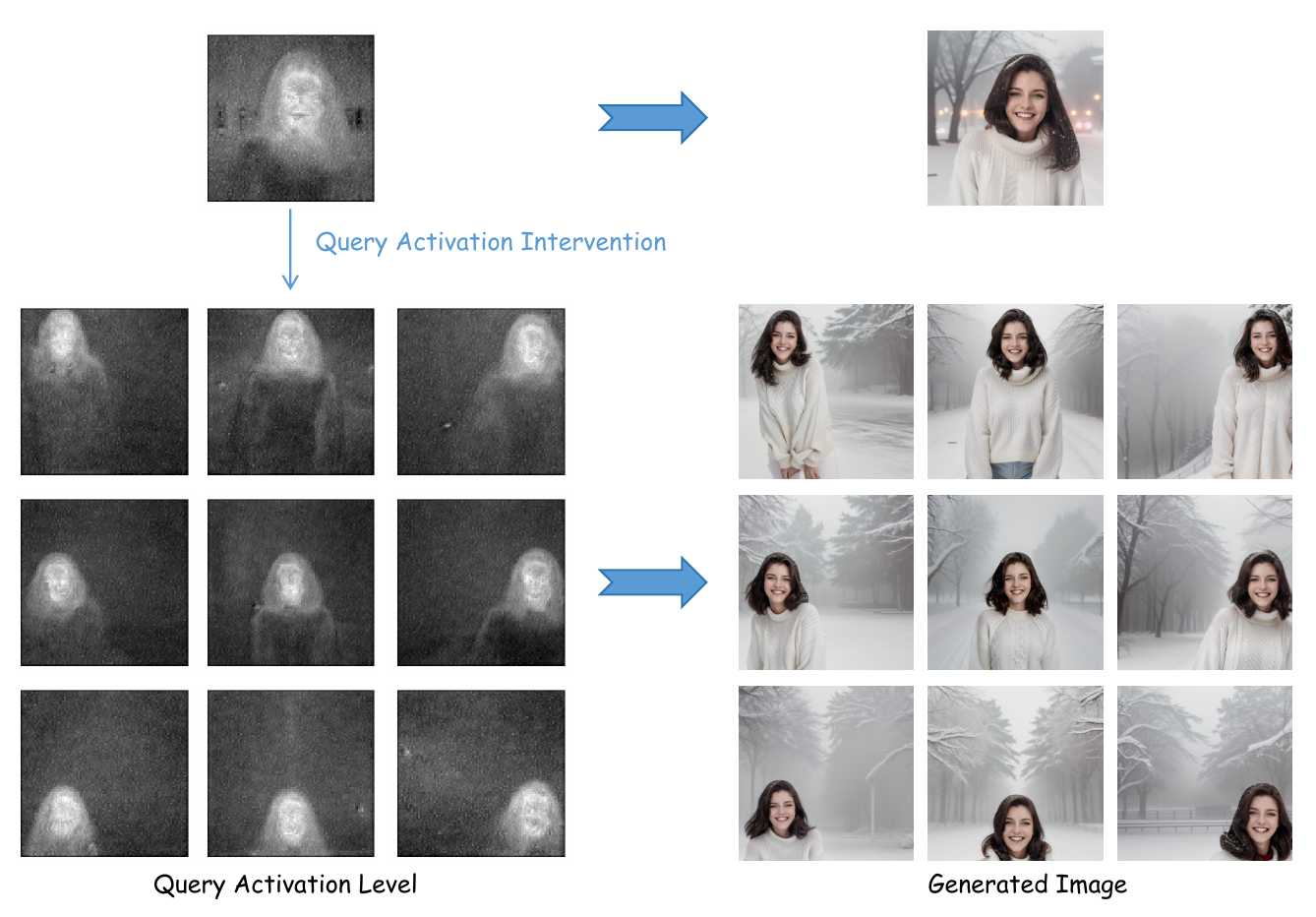}
\caption{The application of layout control.}
\label{layout control}
\end{figure*}

\begin{figure*}[h]
\centering
\includegraphics[width=0.98\textwidth]{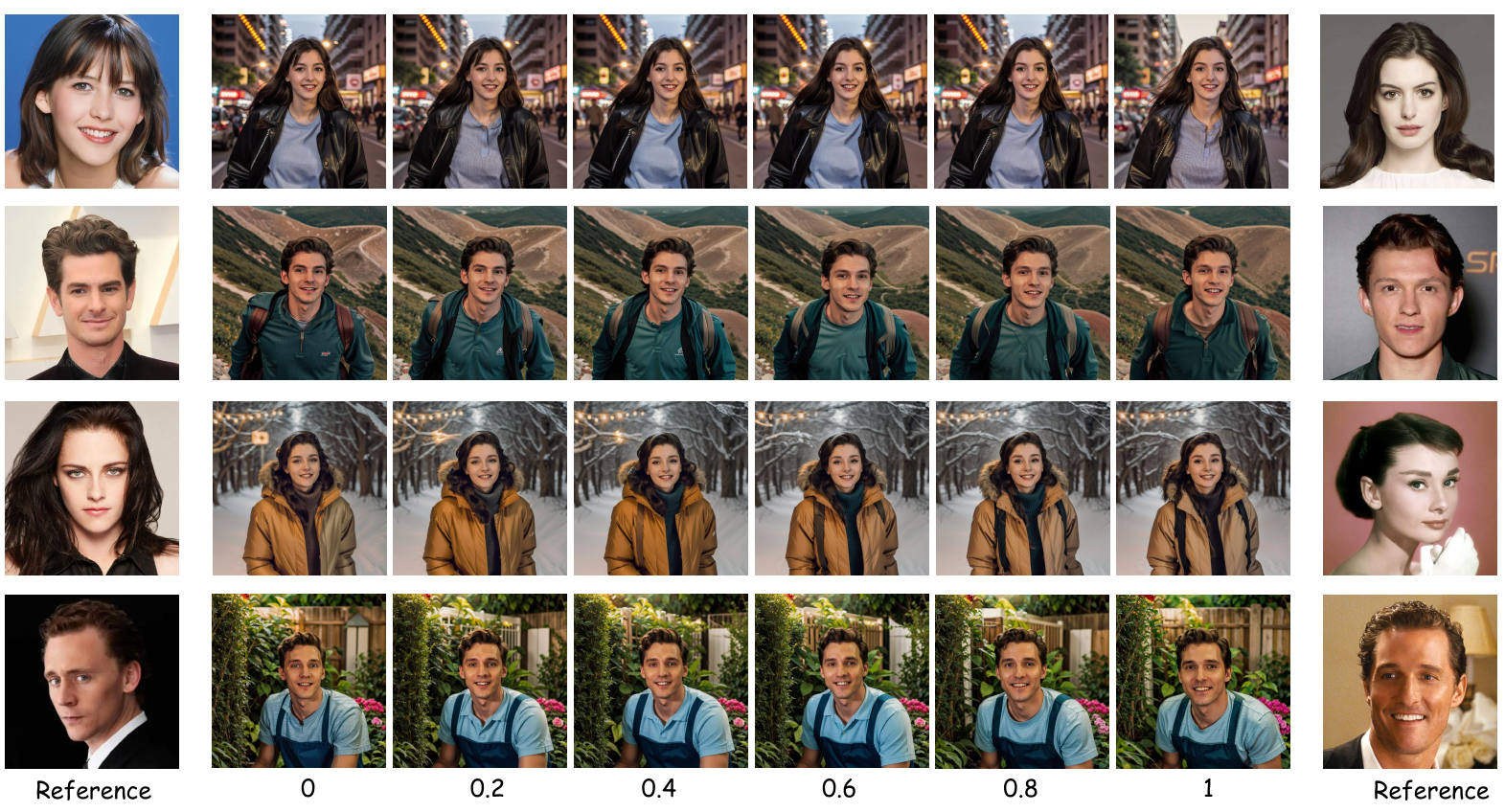}
\caption{The application of ID mixing.}
\label{mix}
\end{figure*}

\begin{figure*}[h]
\centering
\includegraphics[width=0.98\textwidth]{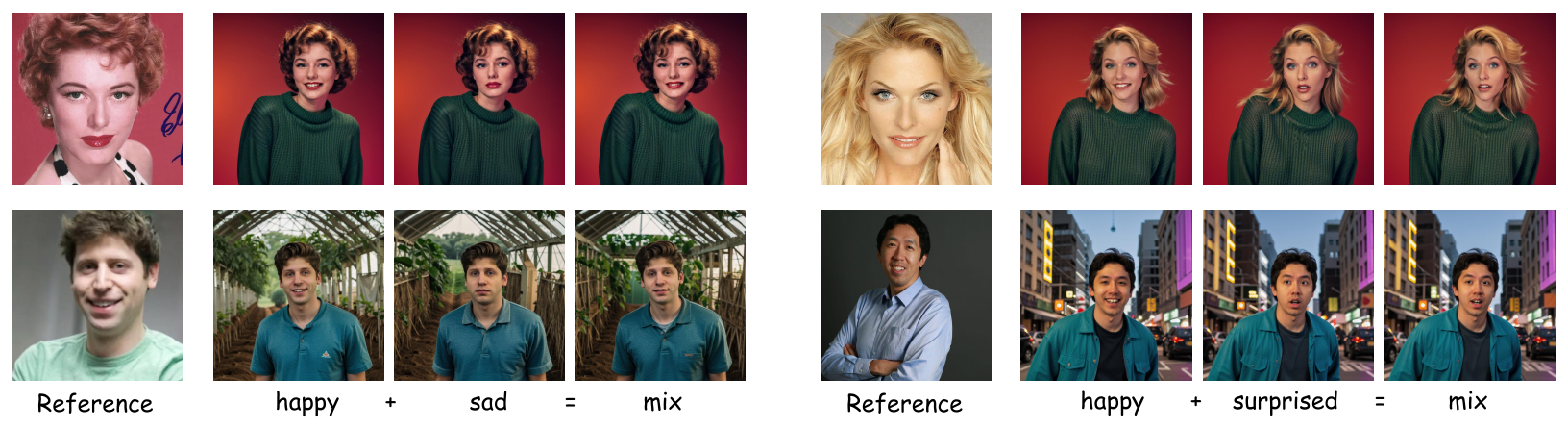}
\caption{The application of complex expression editing. Zoom in for a better view.}
\label{expression}
\end{figure*}

\begin{figure*}[h]
\centering
\includegraphics[width=0.98\textwidth]{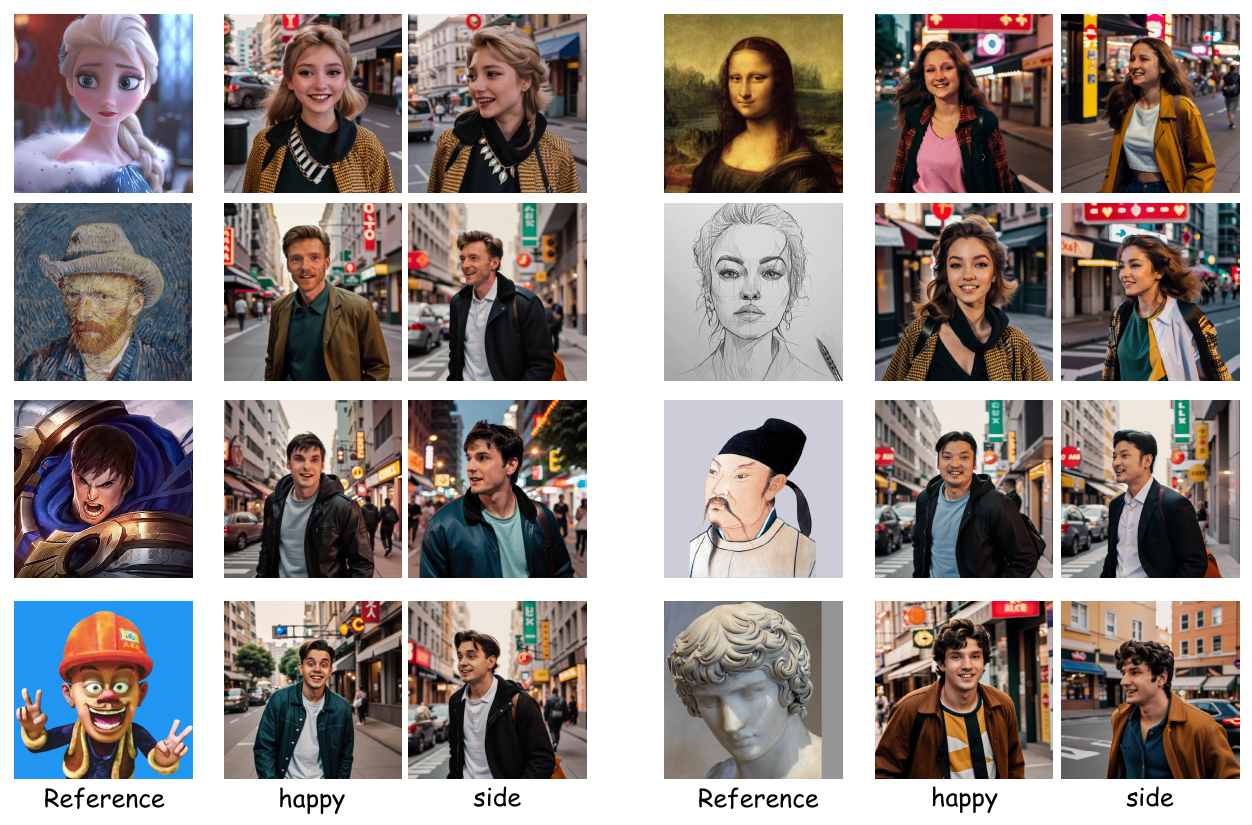}
\caption{The application of transformation from non-photo-realistic domains to photo-realistic ones.}
\label{raw}
\end{figure*}

We provide more applications of our DynamicID, encompassing context decoupling (Fig.~\ref{context}), layout control (Fig.~\ref{layout control}), complex expression editing (Fig.~\ref{expression}), ID mixing (Fig.~\ref{mix}), and transformation from non-photo-realistic domains to photo-realistic ones (Fig.~\ref{raw}).

\begin{table}[h]
\centering
\begin{tabularx}{0.45\textwidth}{l|X}  
\hline
Category & Prompt \\ 
\hline

\multirow{2}{*}{Accessory} 
& wearing headphones \\
& with long yellow hair \\
\hline

\multirow{4}{*}{Clothing} 
& wearing a spacesuit \\
& in a chef outfit \\
& in a doctor's outfit \\
& in a police outfit \\
\hline

\multirow{7}{*}{Background} 
& standing in front of a lake \\
& in the mountains \\
& on the street \\
& in the snow \\
& in the desert \\
& on the sofa \\
& on the beach \\
\hline 

\multirow{5}{*}{Action} 
& reading books \\
& walking on the road \\
& playing the guitars \\
& holding a bottle of red wine \\
& eating lunch \\
\hline 

\multirow{2}{*}{Style} 
& a painting in the style of Ghibli anime \\
& a painting in the style of watercolor \\  
\hline
\end{tabularx}

\caption{Evaluation text prompts are categorized into Clothing, Accessories, Background, Action, and Style, which will be incorporated as part of the final prompt.}
\label{tab:prompta}
\end{table}


\end{document}